\documentclass{article}

\usepackage[preprint]{neurips_2023}

\usepackage[utf8]{inputenc} 
\usepackage[T1]{fontenc}    
\usepackage{hyperref}       
\usepackage{url}            
\usepackage{booktabs}       
\usepackage{amsfonts}       
\usepackage{nicefrac}       
\usepackage{microtype}      
\usepackage{xcolor}         
\usepackage{graphicx}       
\usepackage{hyperref}       

\title{SegmATRon: Embodied Adaptive Semantic Segmentation for Indoor Environment}

\author{%
  Tatiana Zemskova\textsuperscript{\rm 1},
  Margarita Kichik\textsuperscript{\rm 1},
  Dmitry Yudin\textsuperscript{\rm 1, 2},\\
  \textbf{Aleksei Staroverov}\textsuperscript{\rm 1, 2, 3}, \textbf{Aleksandr Panov}\textsuperscript{\rm 1, 2}
   \\
  \textsuperscript{\rm 1}Moscow Institute of Physics and Technology,
  \textsuperscript{\rm 2}Artificial Intelligence Research Institute,
  \\
  \textsuperscript{\rm 3}Federal Research Center for Computer Science and Control of Russian Academy of Sciences \\
  \texttt{zemskova.ts@phystech.edu,  kichik.mg@phystech.edu, yudin.da@mipt.ru,}\\
  \texttt{panov.ai@mipt.ru,  alstar8@yandex.ru} \\
}

\begin{document}

\maketitle

\begin{abstract}
   This paper presents an adaptive transformer model named SegmATRon for embodied image semantic segmentation. Its distinctive feature is the adaptation of model weights during inference on several images using a hybrid multicomponent loss function. We studied this model on datasets collected in the photorealistic Habitat and the synthetic AI2-THOR Simulators. We showed that obtaining additional images using the agent's actions in an indoor environment can improve the quality of semantic segmentation. The code of the proposed approach and datasets are publicly available at \href{https://github.com/wingrune/SegmATRon}{github.com/wingrune/SegmATRon}.
\end{abstract}

\section{Introduction} 
\label{sec:intro}
Embodied Artificial Intelligence involves studying agents that can solve intellectual tasks while interacting with the environment autonomously \citep{pfeifer2004embodied, deitke2022retrospectives}.
This is especially important for modern robots, which must perform reliable scene recognition using onboard sensors (usually cameras) while simultaneously performing navigation or object manipulation tasks \citep{RoomR, partsey2022mapping, staroverov2023skill}.

Recently, embodied methods in object detection \citep{yang2019embodied, kotar2022interactron, wu2022smart, ding2023learning} have appeared, which demonstrate that the information fusion from an image sequence during indoor navigation positively affects the quality of detection. However, the existing embodied approaches do not consider semantic segmentation, another important perception task for intelligent agents.

Inspired by work \citep{kotar2022interactron}, we propose and investigate an adaptive learning method with different action policies for the improvement of semantic segmentation in the Habitat \citep{habitatchallenge2023} and AI2-THOR \citep{kolve2017ai2} indoor environments. 
These environments are among the most popular for researching the problems of interactive perception and navigation of embodied agents.

Our contributions are the following: 

\begin{figure}[t]
  \centering
   \includegraphics[width=0.8\linewidth]{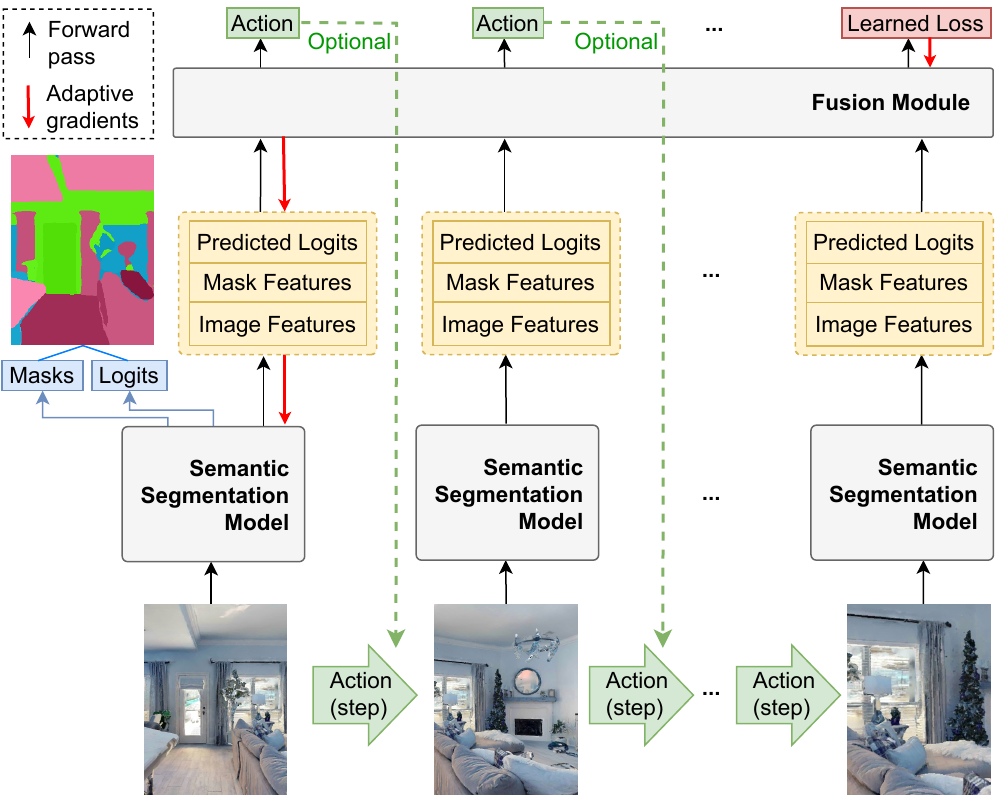}

   \caption{Simplified inference scheme of the proposed SegmATRon approach. Adapting the Semantic Segmentation Model weights during inference on several images is made via learned loss predicted by the Fusion Module to improve the segmentation quality of the first frame. The Transformer-based Fusion Module inputs predicted segment logits, mask features, and image features from the Semantic Segmentation Model. The output of the Fusion module consists of predicted learned loss and, optionally, action. The action can be used to choose the next frame. The Fusion module infers the learned loss when the necessary number of frames is collected.
}
   \label{fig:segmatron_simple}
\end{figure}
\begin{itemize}
     \item We propose a new architecture for an adaptive semantic segmentation neural network called \textit{SegmATRon} (see Figure~\ref{fig:segmatron_simple}). 
     \item We develop a transformer \textit{Fusion module}, which takes image and mask features, predicts semantic logits and masks as inputs and generates output actions that an intelligent agent can perform in the environment to obtain new images. 
     \item We introduce the multicomponent hybrid loss function involving adaptive learned loss, which value is predicted by the SegmATRon. This loss value is then used in the inference to adapt the basic semantic segmentation model. It leads to an increase in the segmentation quality of the first image of the sequence.
     \item To study the quality metrics of embodied semantic segmentation, we create two novel datasets based on the Habitat and AI2-THOR simulators, which contain not only images and masks of semantic segmentation but also a tree of possible actions that an agent can perform from some point in indoor scenes. Thus, we demonstrate the possibility of using our approach in a multi-embodied mode.
\end{itemize}

\section{Related works}
\label{sec:related_works}

\textbf{Image Semantic Segmentation.}
To address the semantic segmentation task, methods based on CNNs and more recent transformer-based approaches have been developed. 

The newest but CNN-based foundation model InternImage \citep{wang2022internimage} and large HRNet-based \citep{WangSCJDZLMTWLX19} methods with attention mechanisms like HRNet+OCR \citep{tao2020hierarchical} and HRNetV2-OCR+PSA \citep{Liu2021PSA} belong to the first category. 

Transformer-based OneFormer \citep{jain2023oneformer} belongs to the second category. It outperforms other state-of-the-art methods, such as Mask2Former \citep{cheng2021mask2former}, k-means Mask Transformer \citep{yu2022kmeans}, and Panoptic-Deeplab \citep{cheng2020panoptic} in solving tasks of semantic, instance and panoptic segmentation. Notably, these achievements are attained without needing to train separately for each task.

Recently, the foundation model Segment Anything (SAM)~\citep{kirillov2023segment} has gained popularity for image segmentation tasks. However, this model doesn't suit semantic segmentation because SAM predicts the segmentation masks in a class-agnostic manner.

\textbf{Video Segmentation.} 
An embodied agent receives information about an environment through a frame sequence. Classical Computer Vision methods, which don't consider camera movement, solve the task of frame sequence segmentation in the scope of Video Segmentation. Recently, densely annotated benchmarks such as CityScapes-VPS \citep{kim2020vps}, VIPSeg \citep{miao2022large}, and VIPOSeg \citep{xu2023video} have appeared, which led to the emergence of video instance segmentation methods. 

TarVIS \citep{athar2023tarvis} is flexible for solving segmentation and detection tasks, MaskFreeVIS \citep{maskfreevis} doesn't use masks for training, DVIS \citep{DVIS}, which implements the decoupling strategy for video instance segmentation, Video-kMaX \citep{shin2023video}, which goal is to bridge the gap between online and offline video segmentation methods --- these and other methods are capable of predicting a category for every pixel of video frames. A distinguishing feature of our methods compared to methods for video segmentation is the adaptive loss function facilitating the model adaptation across different indoor environments without fine-tuning. Furthermore, the mentioned methods require a sequence of frames to be provided, whereas our approach uses only 2 or 4 frames acquired from distinct domains.

\textbf{Embodied Computer Vision.}
Several environments simulating living spaces have been developed for embodied agents, including Habitat \citep{yadav2023habitat} and AI2-THOR \citep{kolve2017ai2}, enabling navigation within the environment and object interactions. A wide range of embodied computer vision methods is present in the field.

The recent work \citep{ding2023learning} proposes to learn a policy for navigation that maximizes the confidence score of a frozen object detector. \citep{yang2019embodied} and \citep{chaudhary2023active} learn to maximize segmentation quality by selecting the next best view based on image features derived from neural network models, whereas \citep{hoseini2022one} demonstrates that a voting system based on four criteria derived from initial viewpoint can improve the object recognition. \citep{wu2022smart}, \citep{liu2022ge}, and \citep{luoae} exploit different policies for push actions to increase the quality of instance segmentation for an embodied agent with gripper. 
However, the existing embodied approaches do not consider semantic segmentation, another important perception task for embodied agents.

Active exploration is crucial in developing embodied agents capable of acting in complex or unfamiliar environments. Examples of such agents include Ask4Help \citep{singh2022ask4help}, which uses human expert hints, and Move to See Better \citep{fang2020better}, which uses multiple frames for fine-tuning during testing. 

Another instance of an active embodied agent is Interactron \citep{kotar2022interactron}, which involves continuous fine-tuning of the detector model during inference. A supervisor is incorporated into the model to adjust the detector's parameters and determine the action policy. The agent navigates through the environment, executing actions from the predetermined set of actions. A notable feature of the Interactron is its adaptive loss function. 

Our work applies a similar approach to address the semantic segmentation task. We introduce a new set of actions and demonstrate that executing just a single additional action is sufficient to enhance segmentation quality. Additionally, we explore the potential to speed up the model's inference time.
The adaptive learned loss function in our method improves the model quality and its ability to generalize to unseen environments. Another strategy for effectively retraining computer vision models in the environment is to collect data based on feedback from the computer vision model. Our method presents the advantage of facilitating adaptation to new domains without necessitating further retraining, along with subsequent inference to improve semantic segmentation quality.

\section{Method}
\label{sec:method}
\begin{figure}[!h]
  \centering
   \includegraphics[width=0.74\linewidth]{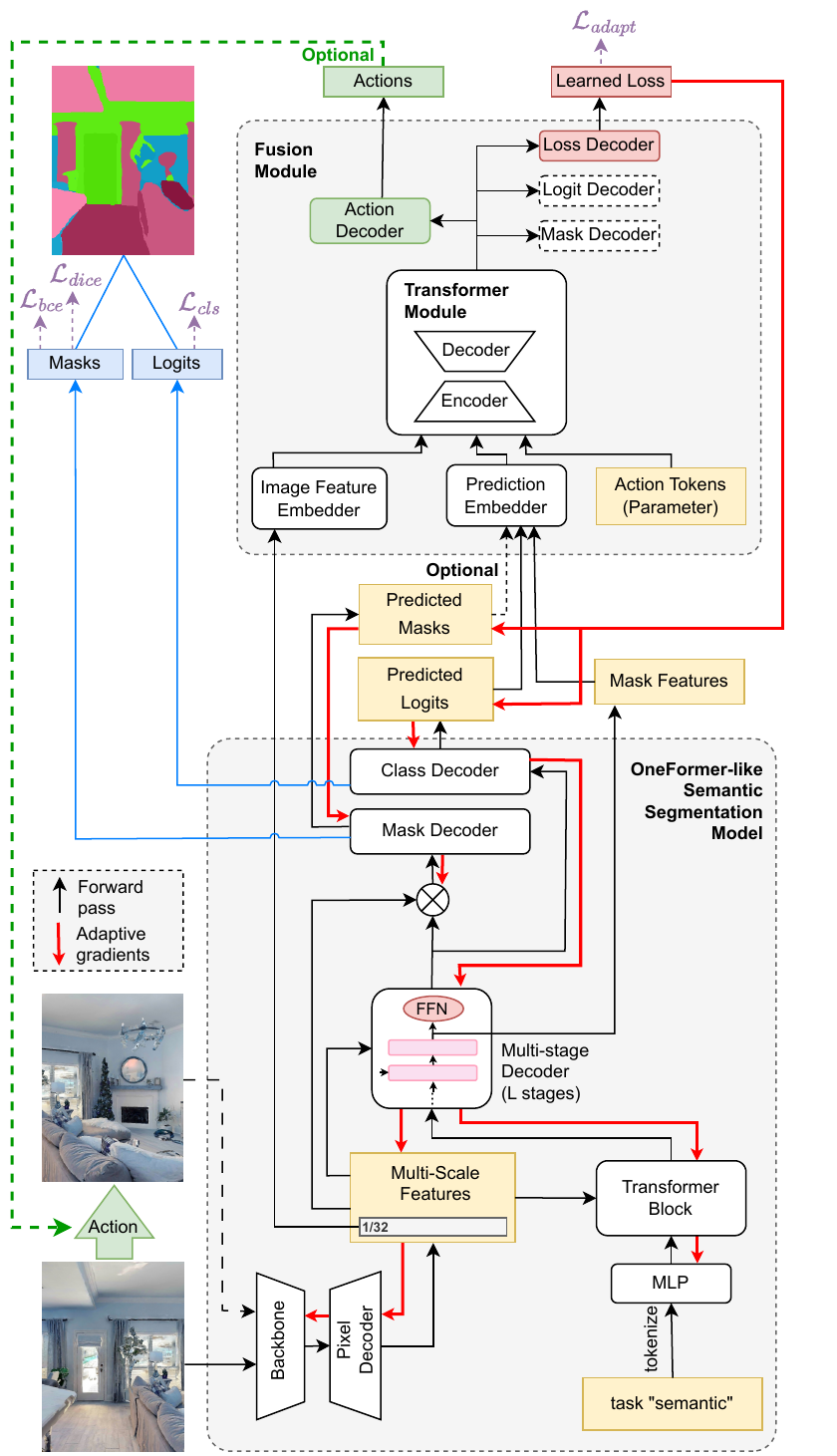}

   \caption{Detailed scheme of the SegmATRon approach.
   It includes two main parts: OneFormer-like Semantic Segmentation Model and Fusion Module.  The Semantic Segmentation Model consists of Image Backbone, Pixel Decoder, MLP Task Encoder, Transformer Block, Multi-Stage Decoder, Mask and Class Decoders. In Ablation Studies, during training, we tried to perform experiments with the freezing of different elements of the Segmentation Model. The Fusion Module aggregates features and predictions of the Segmentation Model and predicts Actions (optional) and Learned Loss for adaptive inference of SegmATRon. The Fusion Module consists of Image Feature and Prediction Embedders, Transformer Module, and Decoders for Action, Loss, Logits, and Masks. The output segmentation result is shown in blue color. Also, the diagram shows how various data are involved in calculating the considered loss functions.
   }
   \label{fig:onecol}
\end{figure}

\textbf{Transformer model.} 
As a segmentation model (see Figure~\ref{fig:onecol}), we consider the modification of OneFormer~\citep{jain2023oneformer}, which is one of the state-of-the-art methods for semantic segmentation. The off-the-shelf OneFormer uses a single frame to make predictions of masks and labels representing a baseline approach for comparison with our SegmATRon model. 

Following the idea of Interactron~\citep{kotar2022interactron}, we choose a Transformer model to combine predictions and image features from several frames to predict the loss for the adaptive backward pass. As the Transformer Model, we use the DETR Transformer Decoder \citep{carion2020end}.

The Fusion Module (see Fig.\ref{fig:onecol}) takes as an input the $1/32$ feature map from the Multi-scale Pixel Decoder of OneFormer, predicted logits of mask classification and masks features that are represented by the input of the last FFN layer of the last stage of Multi-stage Decoder of OneFormer. This input is mapped to the dimension of the Transformer module by corresponding embedders. 

We change the architecture of the Prediction Embedder in the Fusion Module compared to the Fusion Module provided by the authors of Interactron~\citep{kotar2022interactron}. We replace a linear layer with an MLP and consider only the mask features, whereas the authors of Interactron~\citep{kotar2022interactron} use predicted boxes and their features as input to the Prediction Embedder of the Fusion Module. The choice of the prediction embedding method is described in detail in the "Ablation studies" Section. The rest of the Fusion Module rests as introduced in the original work ~\citep{kotar2022interactron}. Therefore our Fusion Module contains MLP decoders for the learned loss, masks, logits, and actions. However, in our experiments, we use only the learned loss output.  

During training, the parameters $\phi$ of the Fusion Module are updated by the ground-truth loss computed from the segmentation annotation and predictions made by OneFormer after the backpropagation of adaptive gradients. Then, the parameters of the OneFormer model are optimized to reduce the ground-truth loss with adapted weights. During inference, there is no ground truth, and the parameters of the OneFormer model are updated by the learned loss predicted by the Fusion Module.

\textbf{Adaptive Learning.} The key idea of adaptive semantic segmentation is the adaptation of model weights during inference on several images using a hybrid multicomponent loss function with adaptive learned part
$\mathcal{L}_{adapt}(\phi, \theta, \mathbf{F})$.
The loss function is parameterized by Fusion Module parameters $\phi$ and depends on parameters $\theta$ of a segmentation model and a sequence of frames $\mathbf{F}$. The goal during the training process is to minimize the  multicomponent loss  $\mathcal{L}_{segm}(\theta, \mathbf{F})$ over all ground-truth sequences $\mathbf{R}_{all}$, where the parameters $\theta$ are updated by backpropagation through adaptive gradients: 
\begin{equation}
 \min_{\theta, \phi}{\sum_{\mathbf{F} \in \mathbf{R}_{all}} \mathcal{L}_{segm}(\theta - \alpha\nabla_{\theta}\mathcal{L}_{adapt}(\phi, \theta, \mathbf{F}), \mathbf{F})}.
  \label{eq:adaptive-learning}
\end{equation}

We use OneFormer segmentation loss \citep{jain2023oneformer} without considering the contrastive loss term. In the original OneFormer method, the contrastive loss function is designed to guide the learning of object queries which should capture the difference between segmentation types and categories of image collection preset in a mini-batch. In our method, we consider adjacent frames. Therefore, we don't expect them to have completely orthogonal object queries. Additionally, in the design of the SegmATRon model, only the semantic segmentation task is considered. Thus,
\begin{equation}
 \mathcal{L}_{segm} = \lambda_{cls} \mathcal{L}_{cls} + \lambda_{bce} \mathcal{L}_{bce} + \lambda_{dice} \mathcal{L}_{dice},
  \label{eq:segmentation-loss}
\end{equation}
where, $\mathcal{L}_{cls}$ -- cross-entropy loss for class prediction, binary cross-entropy ($\mathcal{L}_{bce}$) and dice loss ($\mathcal{L}_{dice}$) are controlling 
mask predictions. We use the set of hyper-parameters proposed in the OneFormer for segmentation loss $\lambda_{cls} = 2$, $\lambda_{bce} = 5$ and  $\lambda_{dice} = 5$, $\lambda_{cls}$ is set to $0.1$ for the no-object prediction. 


\begin{table*}[t]
  \centering
  \caption{Comparison of SegmATRon method and the state-of-the-art OneFormer model on the Habitat and AI2-THOR datasets with 150 categories.
  In parentheses, here and below, we show the relative increment of the quality metric compared to the baseline.
  }
  \label{tab:results}
  \scriptsize
  \begin{tabular}{@{}p{2.2cm}@{}p{1.35cm}@{}p{1.5cm}@{}p{1cm}@{}p{1.2cm}@{}p{1.65cm}@{}p{1.65cm}@{}p{1.65cm}@{}p{1.65cm}@{}}
    \hline
    Method & Adaptation on Inference & Action Policy & Training Dataset & Validation Dataset & $mIoU$, \%  & 
      $fwIoU$, \%  & $mACC$, \% &  $pACC$, \% \\
    \hline\hline
    OneFormer & no & Single Frame & Habitat & Habitat & $25.1 $ & $70.7 $ & $34.7 $ & $80.4$ \\
    SegmATRon (1 Step) & \textbf{yes} & Random & Habitat  & Habitat & $\mathbf{26.7}$ $(+6.4\%)$& $\mathbf{71.1}$ $(+0.6\%)$& $\mathbf{36.7}$  $(+5.8\%)$ & $\mathbf{80.6}$ $(+0.2\%)$ \\
    SegmATRon (1 Step)  & no & Random & Habitat & Habitat &$26.6$ $(+5.9\%)$& $70.9$ $(+0.3\%)$& $36.6$ $(+5.5\%)$ & $80.4$ $(+0.0\%)$ \\
    \hline
    OneFormer & no & Single Frame & Habitat & AI2-THOR &$30.7$ & $45.4$ & $42.2$ & $60.5$ \\
    SegmATRon (1 Step) & \textbf{yes} & Random & Habitat & AI2-THOR & $\mathbf{32.0}$ $(+4.2\%)$& $\mathbf{49.0}$ $(+7.9\%)$& $\mathbf{44.1}$ $(+4.5\%)$& $\mathbf{63.9}$ $(+5.6\%)$ \\
    SegmATRon (1 Step) & no & Random & Habitat & AI2-THOR& $31.6$  $(+2.9\%)$& $48.5$ $(+6.8\%)$& $43.7$ $(+3.6\%)$& $63.5$ $(+4.9\%)$\\
    \hline
  \end{tabular}
\end{table*}

\section{Datasets for Adaptive Learning in Indoor Environment}
\label{sec:dataset}
\textbf{Habitat environment.} OneFormer model was pre-trained using 500K images, collected in random navigable points of train scenes from Habitat Matterport3D semantics (HM3DSem) v0.1 dataset \citep{yadav2023habitat} with 150 categories from ADE20k dataset \citep{zhou2019ade20k} to enrich the diversity of scene semantics compared to the original 40 Matterport3D categories \citep{Matterport3D}. We use their ground truth masks for overlapping categories between Matterport3D and ADE20k datasets. Masks for categories absent in the Matterport3D dataset are obtained with pseudo-labeling using an efficient model SegFormer \citep{xie2021segformer} pre-trained on the ADE20k dataset.  

To train our SegmATRon models, we collected a dataset of $1160$ points in train scenes of HM3DSem v0.2 \citep{yadav2023habitat} with possible additional points of view. A validation dataset of 144 points was collected from validation scenes of HM3DSem v0.2. For the train and the validation datasets, we considered all possible combinations of 4 additional frames obtained with the following agent actions: turn left, turn right, look up, look down, and move backward. The last action corresponds to observing a scene from a more distant point of view. All rotations are made by 30\textdegree.

Since HM3DSem v0.2 contains two sets of categories for semantic segmentation annotation, the first set contains 40 Matterport3D categories \citep{Matterport3D}. The second set contains a rich semantic with 1624 categories. We decided to leverage this large set of categories and map them into 150 ADE20k categories, which allowed us to get ground truth semantics without pseudo-labeling. For matching categories, we left their original names. Object categories having supercategories in the ADE20K dataset were assigned to their supercategory (e.g., wine bottle - bottle, apple - food, solid food). Small objects with a familiar location in scenes were assigned to their location (e.g., pen-table). Small objects that do not have a fixed location were categorized as unlabeled (e.g., sponge - unlabeled). 

The frame rendering parameters correspond to the Habitat Navigation Challenge 2023 \citep{habitatchallenge2023} configuration. In particular, the image size was fixed to $640\times480$, horizontal field of view angle was equal to 42\textdegree.

\textbf{AI2-THOR environment.}
To test the domain adaptation ability of our models, we collected a test dataset of 100 points in the test scenes of the iTHOR synthetic environment \citep{kolve2017ai2} using the same set of actions and the same render settings as for the Habitat environment. As the categories set in the AI2-THOR simulator differ from the environment in the Habitat simulator, we considered only 45 intersecting categories from the available 125 categories in the iTHOR scenes.

\begin{figure*}[!htb]
  
   \centering
   \includegraphics[width=0.8\textwidth]{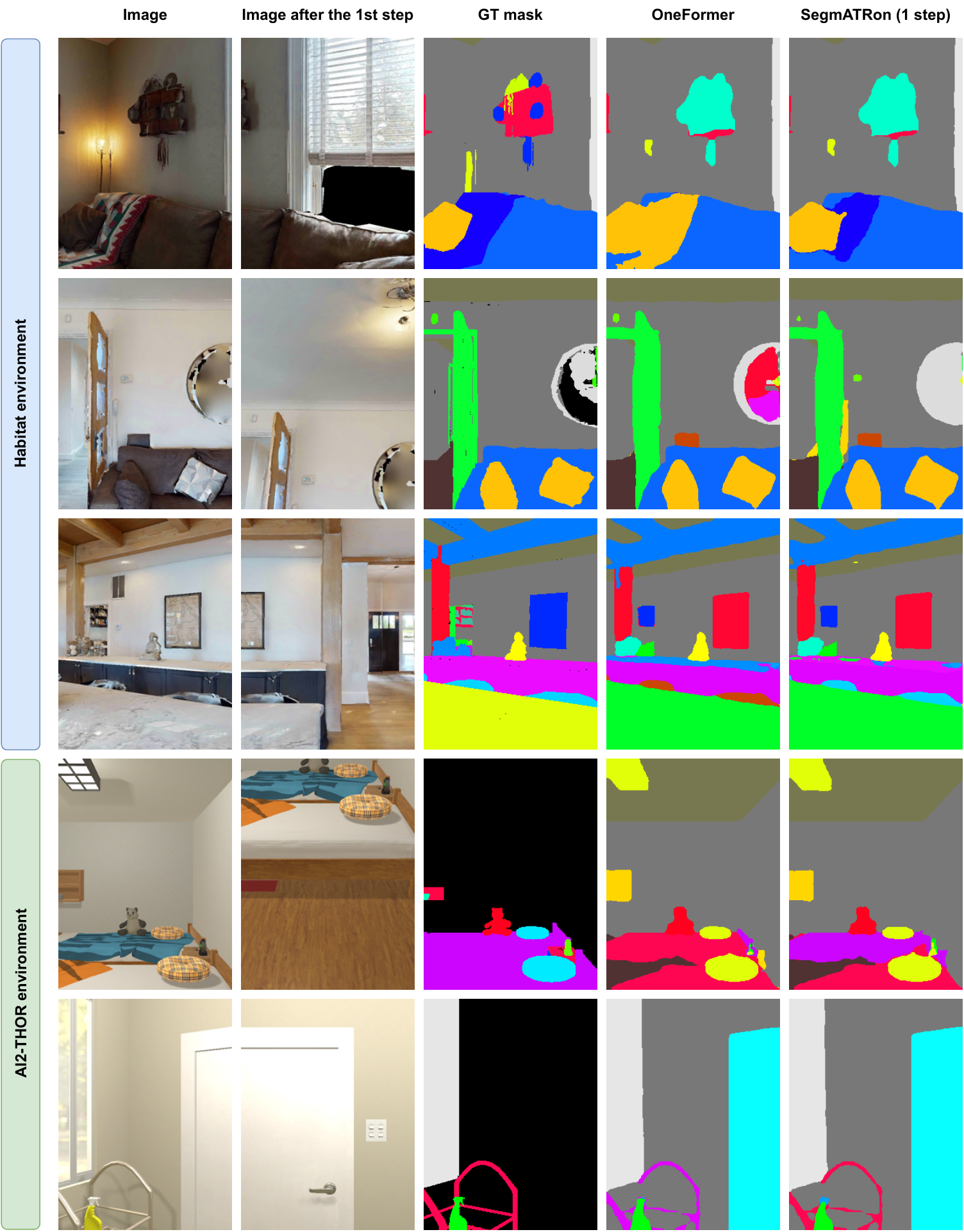}
   \caption{
  Visualized segmentation results on Habitat and AI2-THOR validation sets. The columns left-to-right refer to the input image, the image, received after one step, the ground truth, the outputs of the OneFormer model and our SegmATRon.
   }
   \label{fig:qualitative-results}
\end{figure*}

\section{Experiments}

\textbf{Training setup.}
We train neural network models on a server with 2$\times$Nvidia Tesla V100 GPU. We pre-train the OneFormer model with Swin-L backbone, crop size 640$\times$640, and batch size equal to 4 following the original training procedure of the OneFormer model~\citep{jain2023oneformer}. The weights are initialized by the OneFormer model trained on ADE20k \citep{zhou2017scene}. To train SegmATRon as well as Oneformer we follow a training procedure described by authors of Interactron \citep{kotar2022interactron}, but we reduce the epoch number to 50 due to the fast convergence of the segmentation model. For SegmATRon (4 Steps) models, we increase the number of epochs to 100.

We trained the models using Adam optimizer with $\beta_1 = 0.9, \beta_2 = 0.999$, gradient clipping with a max norm of 1 and batch size of 16. The learning rate for the segmentation model was set to $10^{-5}$, and the learning rate for the Fusion module was equal to $10^{-4}$. For each model design, we run the training process once. During the training process of SegmATRon, we resize input images to $320\times240$ resolution and pad the image to have a square shape of $320\times320$.

After training for 50 (100) epochs, we choose checkpoints with the best $fwIoU$ value on the validation dataset. We report standard metrics for semantic segmentation \citep{jain2023oneformer}: mean Intersection over union ($mIoU$), frequency-weighted Intersection over union ($fwIoU$), mean pixel accuracy ($mACC$) and pixel accuracy ($pACC)$.

\textbf{OneFormer as Single Frame baseline.}
To distinguish the role of the adaptive learned loss function from the role of fine-tuning the segmentation model, we experimented with fine-tuning the OneFormer model (Swin-L backbone) without the Fusion Module, following the segmentation model training procedure in the SegmATron architecture.

\textbf{Results.}
The SegmATRon with Random rotation action policy significantly outperforms the baseline OneFormer approach (see Table \ref{tab:results}) both on the validation dataset collected in the Habitat environment and the test dataset collected in the different domain of AI2-THOR environment. Since the SegmATRon approach requires the backpropagation of adaptive gradients during inference, more computing resources are needed for this method. We show that the learned hybrid multicomponent loss function not only improves the convergence of the segmentation model but also increases the segmentation quality during inference via adaptive gradients.

Figure \ref{fig:qualitative-results} shows the visualized results of SegmATRon compared to the OneFormer baseline method under different scenes from Habitat and AI2-THOR simulators. The SegmATRon model helps to correctly predict the object masks located in the corners or on the sides of initial frames. In the first image, the SegmATRon correctly segments a blanket. The second image demonstrates the improvement of mirror segmentation. In the third image, the SegmATRon is capable to recognize stools while the OneFormer baseline recognizes chairs. It's worth noting that the misclassification between stools and chairs is a common mistake in the Object Navigation task for embodied agents.

In the last two images in the ground truth masks it is able to see black background. This is a distinctive characteristic of the data compiled using AI2-THOR, which includes the "Unlabelled" category. In the fourth image, SegmATRon more accurately identifies the mask of the bed. In the fifth image, it correctly classifies the chair.

\begin{table*}[!htb]
  \centering
  \caption{Models used in the "Ablation Studies" Section.}
  \label{tab:segmatrons}
  \scriptsize
  \begin{tabular}{@{}p{2cm}p{1.9cm}p{1.9cm}p{1.9cm}p{2cm}p{2cm}@{}}
    \hline
    Method & Backbone & MLP Task Encoder & Pixel Decoder & Transformer Block & Multi-stage Decoder \\
    \hline\hline
    OneFormer& finetuned & finetuned & finetuned & finetuned & finetuned \\
    \hline
    OneFormer$^{\dag}$& frozen & frozen & frozen & finetuned & finetuned  \\
    \hline
    SegmATRon Tiny & frozen & frozen & frozen & finetuned, adaptive & finetuned, adaptive  \\
    \hline
    SegmATRon Small & finetuned & finetuned & finetuned, adaptive & finetuned, adaptive & finetuned, adaptive  \\
    \hline
    SegmATRon  & finetuned, adaptive & finetuned, adaptive & finetuned, adaptive & finetuned, adaptive & finetuned, adaptive   \\
    \hline
  \end{tabular}
\end{table*}

\begin{table*}[!htb]
  \centering
  \caption{Ablation study. Comparison of different Prediction Embedder types.}
  \label{tab:embeddings}
  \scriptsize
  \begin{tabular}{@{}p{2.1cm}p{2.1cm}p{1.8cm}p{1.8cm}p{1.8cm}p{1.8cm}@{}}
    \hline
    Method & Prediction Embedder & $mIoU$, \%  & 
      $fwIoU$, \%  & $mACC$, \% &  $pACC$, \% \\
    \hline\hline
    OneFormer & - & $25.1 $ & $70.7 $ & $34.7 $ & $80.4$ \\
    SegmATRon (1 Step) & Vanilla & $25.8$ $(+2.8\%)$ & $\mathbf{71.2}$ $(+0.7\%)$ & $34.9$ $(+0.6\%)$ & $\mathbf{80.8}$ $(+0.5\%)$ \\
    SegmATRon (1 Step) & MLP & $\mathbf{26.7}$ $(+6.4\%)$& $71.1$ $(+0.6\%)$& $\mathbf{36.7}$ $(+5.8\%)$&$80.6$ $(+0.2\%)$ \\
    \hline
  \end{tabular}
\end{table*}

\begin{figure*}[!htb]
   \centering
   \includegraphics[width=1\textwidth]{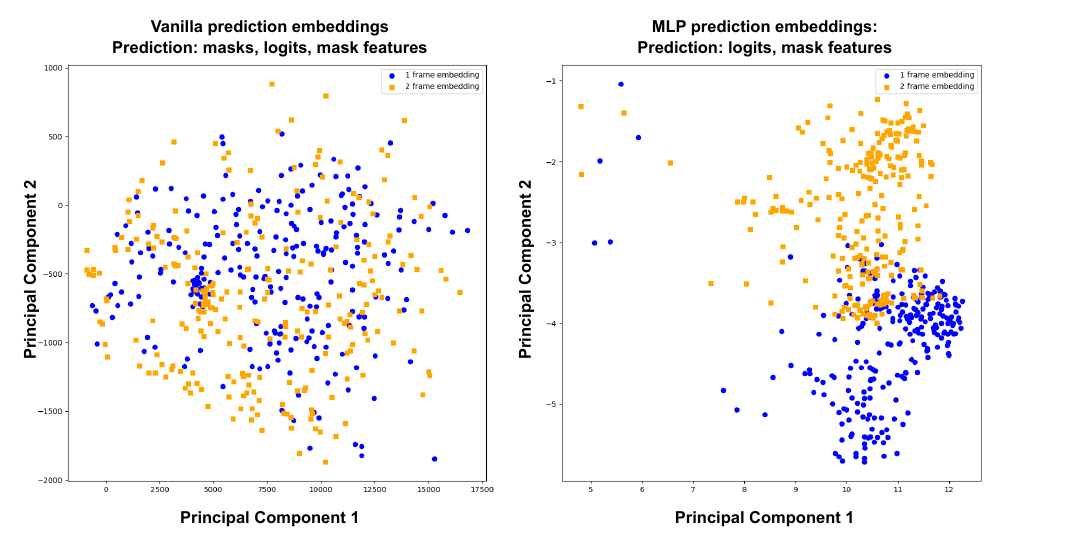}
   \caption{
An illustration demonstrating how the choice of the Prediction Embedder method can impact the input for the Transformer Module of the Fusion model. The figure shows first the two Principal Components of the Prediction Embedder output. \textbf{Left}: a vanilla approach. Prediction embedding is formed by a Linear Layer with predicted masks, logits, and mask features as input. Two different frames have close prediction embeddings. \textbf{Right}: the Principal Components Analysis results for the output of SegmATRon Prediction Embedder for the same Habitat Validation dataset point. The prediction embeddings for different frames are well separated.
}
\label{fig:embeddings}
\end{figure*}

\section{Ablation studies}
\label{sec:ablation}
We analyze SegmATRon’s components through a series
of ablation studies with different models (see Table \ref{tab:segmatrons}).

\textbf{Prediction embeddings.} We tested two different architectures for the Prediction Embedder of the Fusion Module. As one can see from Table \ref{tab:embeddings} the MLP which takes as an input the mask features and predicted class logits gives a significant gain in segmentation quality for $mIoU$ and $mACC$ metrics compared to the vanilla approach. The vanilla method for prediction embedding consists of a Linear Layer that takes predicted masks, logits, and mask features as input. The $fwIoU$ and $pACC$ metrics are slightly lower for MLP Prediction Embedder but the increase of $mIoU$ and $mACC$ metrics is more significant. Figure \ref{fig:embeddings} shows the results of Principal Component Analysis for the output of the Prediction Embedder block for these two approaches for the same data point of the Habitat Validation dataset. The MLP Prediction Embedder gives embeddings for different frames that are well separated.

\begin{table*}[!htb]
  \centering
  \caption{Ablation study. The number of steps (additional frames).}
  \label{tab:frames}
  \scriptsize
   \begin{tabular}{@{}p{2.5cm}@{}p{2cm}@{}p{2.4cm}@{}p{2.2cm}@{}p{2.2cm}@{}p{2.4cm}@{}}
    \hline
    Method & Number of steps & $mIoU$, \%  & 
      $fwIoU$, \%  & $mACC$, \% &  $pACC$, \% \\
    \hline\hline
    OneFormer & -  & $25.1 $ & $70.7 $ & $34.7 $ & $80.4$ \\
    \hline
    SegmATRon Tiny & 1  & $25.7$ $(+2.4\%)$ & $69.0$ $(-2.4\%)$& $34.9$ $(+0.6\%)$ & $79.4$ $(-1.2\%)$\\
    SegmATRon Tiny  & 4  & $\mathbf{26.4}$ $(+5.2\%)$& $\mathbf{69.9}$ $(-1.1\%)$& $\mathbf{35.5}$ $(+2.3\%)$ & $\mathbf{80.4}$ $(+0.0\%)$ \\
    \hline
    SegmATRon Small & 1 & $26.6$ $(+6.0\%)$ & $70.9$ $(+0.3\%)$ & $\mathbf{35.5}$ $(+2.2\%)$ & $80.4$ $(+0.0\%)$\\
    SegmATRon Small & 4 & $\mathbf{27.1}$ $(+8.0\%)$& $\mathbf{71.2}$ $(+0.7\%)$& $34.9$ $(+0.6\%)$&$\mathbf{80.8}$ $(+0.5\%)$ \\
    \hline
    SegmATRon & 1 & $\mathbf{26.7}$ $(+6.4\%)$ & $\mathbf{71.1}$ $(+0.6\%)$ & $\mathbf{36.7}$ $(+5.8\%)$ &$80.6$ $(+0.2\%)$ \\
    SegmATRon & 4 & $25.3$ $(+0.8\%)$& $71.0$ $(+0.4\%)$& $34.7$ $(+0.6\%)$& $\mathbf{80.9}$ $(+0.6\%)$\\
    \hline
  \end{tabular}
\end{table*}

\begin{table*}[!htb]
  \centering
  \caption{Ablation study. Computational resources.}
  \label{tab:resolution}
  \scriptsize
  \begin{tabular}{@{}p{2.8cm}@{}p{1.2cm}@{}p{1.8cm}@{}p{1.8cm}@{}p{1.8cm}@{}p{1.8cm}@{}p{2.75cm}@{}}
    \hline
    Method & Resolution & $mIoU$, \%  & 
      $fwIoU$, \%  & $mACC$, \% &  $pACC$, \% & Inference GPU Memory, Mb\\
    \hline\hline
    OneFormer$^{\dag}$& 320 & $25.8 $ & $70.0 $ & $35.0 $ & $80.4$ & $2692$\\
    OneFormer$^{\dag}$& 640 & $\mathbf{27.0} $ $(+4.7\%)$ & $69.1$ $(-1.3\%)$& $\mathbf{36.0}$ $(+2.9\%)$& $79.5$ $(-1.1\%)$ & $4304$\\
    SegmATRon Tiny (4 Steps) & 320 & $26.4$ $(+2.3\%)$& $\mathbf{69.9}$ $(-0.1\%)$& $35.5$ $(+1.4\%)$ & $\mathbf{80.4}$ $(+0.0\%)$ & $6288$\\
    \hline
  \end{tabular}
\end{table*}

\begin{table*}[!htb]
  \centering
  \caption{Ablation study. Policy optimization.}
  \label{tab:policy}
  \scriptsize
  \begin{tabular}{@{}p{5cm}@{}p{1.2cm}@{}p{1.8cm}@{}p{1.8cm}@{}p{1.8cm}@{}p{1.8cm}@{}}
    \hline
    Method & Policy & $mIoU$, \%  & 
      $fwIoU$, \%  & $mACC$, \% &  $pACC$, \% \\
    \hline\hline
    SegmATRon Tiny (4 Steps) (GPT Fusion module) & Random  & $25.7$ & $69.6$ & $34.8$ & $80.1$ \\
    SegmATRon Tiny (4 Steps) (GPT Fusion module) & Best loss & $\mathbf{26.1}$ $(+1.5\%)$& $\mathbf{69.8}$ $(+0.3\%)$& $\mathbf{34.8}$ $(+0.0\%)$ & $\mathbf{80.4}$ $(+0.4\%)$ \\
    \hline
  \end{tabular}
\end{table*}

\textbf{SegmATRon Small and SegmATRon Tiny}. We conduct several ablation studies on the light versions of SegmATRon. These models were trained with the same optimizer parameters and the batch size was equal to $1$. In these experiments, we used the Vanilla architecture for the Prediction Embedder.

SegmATRon Small differs from the full version of SegmATRon by the set of segmentation model parameters that are updated by the adaptive gradient during the backpropagation of the 
learned loss predicted by the Fusion Module. In the SegmATRon Small setup, the adaptive gradients are computed only for parameters $\theta_{head}$ of a semantic segmentation model head which consists of Pixel Decoder and Transformer Block and Multistage Decoder. 

SegmATRon Tiny is a computationally efficient version of SegmATRon. The adaptive gradients are computed only for the Multistage Decoder and Transformer Block of the segmentation model. To compensate for the large effect of segmentation model fine-tuning during the training process of SegmATRon we keep its Backbone, Pixel Decoder, and Task MLP frozen during the training of SegmATRon Tiny. To ensure a fair comparison we also conducted an experiment of training the OneFormer model with the frozen Backbone, Pixel Decoder, and Task MLP which is noted OneFormer$^{\dag}$.

A summary of models used in the Ablation Studies is presented in Table \ref{tab:segmatrons}.

\textbf{Number of Steps (Additional Frames).} We study the influence of the frame number used for the prediction of the 
learned loss function. As one can see from Table \ref{tab:frames} the use of $4$ additional frames instead of $1$ improves the performance of both SegmATRon Tiny and SegmATRon Small models. However, this effect is not observed for the full version of SegmATRon. Therefore, additional experiments are necessary to find an optimal number of additional frames.

\textbf{Computational resources.} The gradient computation during the inference time needs additional computational resources. We compare the impact of increasing the resolution of input images for the Oneformer$^{\dag}$ on both the segmentation quality metrics and GPU memory necessary for the inference process (see Table \ref{tab:resolution}). We observe that the larger input resolution of $640 \times 640$ does not improve the $fwIoU$ and $pACC$ metrics but improves the $mIoU$ and $mACC$ metrics while demanding 60\% more GPU memory. The SegmATRon Tiny (4 Steps) needs 46\% more GPU memory during the inference than the OneFormer model with a resolution of $640 \times 640$ but is capable to improve the $mIoU$ and $mACC$ metrics without worsening the $fwIoU$ and $pACC$ metrics. 

\textbf{Policy Optimization.} Finally, we study the optimization of the policy of choosing the next frame in the sequence. We adopt the approach proposed by the authors~\citep{kotar2022interactron}. Since in this setup, the Fusion Module needs to work with different lengths of sequences we replace the DETR Transformer module with the GPT module.  During the training process, the SegmATRon Tiny gradually explores possible trajectories and learns to predict the best path from the observed. The path is considered the best if it gives the smallest ground truth loss. As one can see from Table \ref{tab:policy} this approach for policy optimization improves the segmentation quality of SegmATRon Tiny.

\section{Conclusion}
\label{sec:conclusion}
Our results show that the semantic segmentation quality benefits from the mechanism of multicomponent loss learning which allows us to use an additional point of view.
We have also demonstrated that the action strategy has a significant impact on the result, while further research on the number of actions and their automatic learning is reasonable.

As a limitation of the proposed approach, we can highlight the difficulty of scaling the approach to more than 4 steps. In this case, the need for video memory increases significantly. Another limitation is the small number of existing datasets for training and testing embodied segmentation methods.

A future perspective for the SegmATRon approach would be action policy optimization via Reinforcement Learning based on segmentation loss, which we are currently working on.
Other promising future directions are the study of other basic semantic segmentation models as part of the proposed approach, as well as its application to solve the problem of instance segmentation.

\bibliographystyle{plainnat}
\bibliography{SegmATRon}

\newpage
\appendix

\section{Appendix: Visualization of SegmATRon Results on Test Images}

\begin{figure*}[!h]
   \centering
   \includegraphics[width=0.96\textwidth]{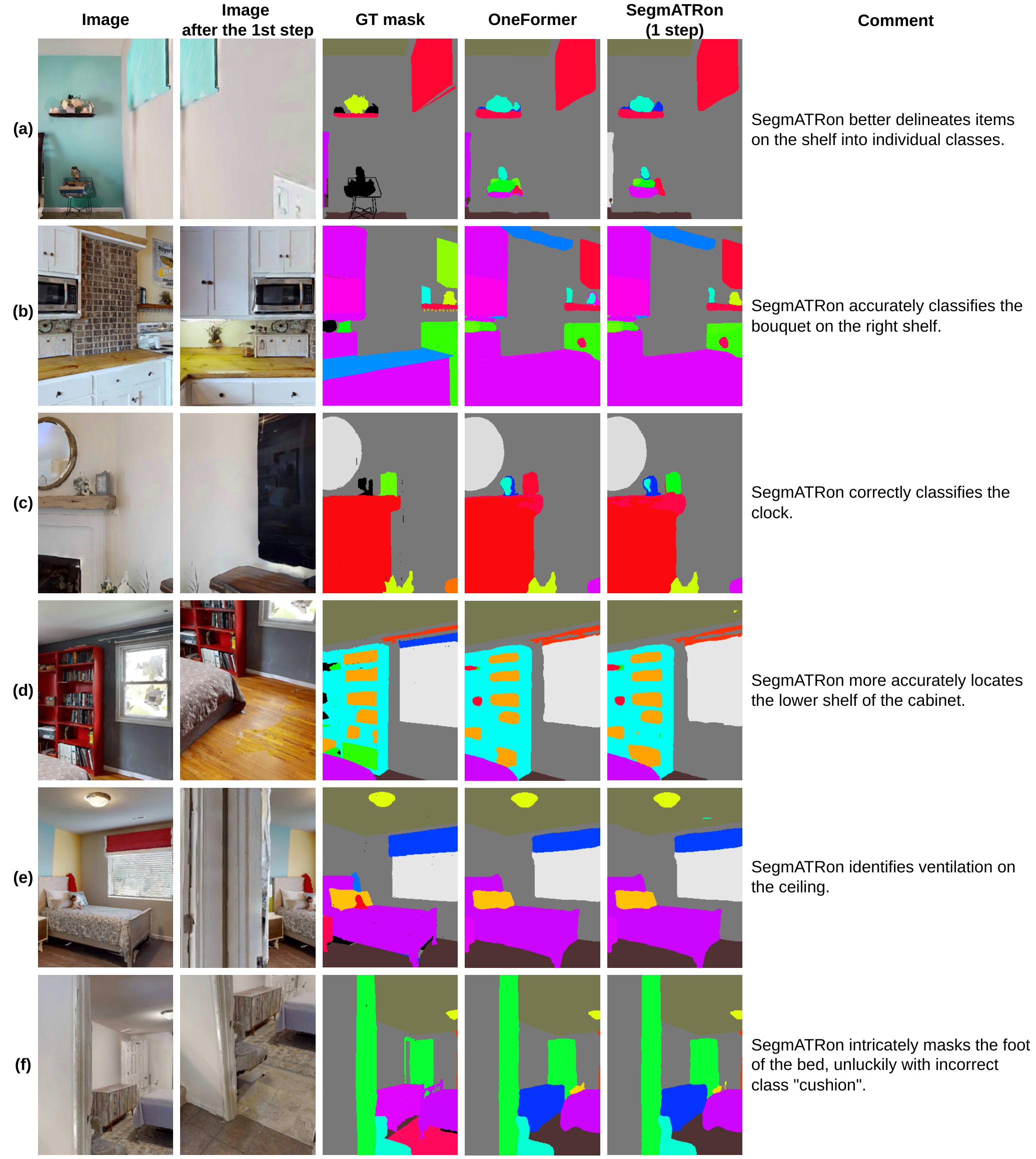}
   \caption{
  Visualized segmentation results on Habitat validation set. The columns left-to-right refer to the input image, the image, received after one step, the ground truth, the outputs of the OneFormer model and our SegmATRon.
   }
   \label{fig:qualitative-results_hab_supplementary}
\end{figure*}

Figure \ref{fig:qualitative-results_hab_supplementary} shows more segmentation results of SegmATRon compared to Oneformer on the images rendered with Habitat. In the provided examples, SegmATRon more accurately identifies the object class compared to OneFormer (images (a)-(c)), achieves greater precision (though not always correctness) in delineating object masks (images (d) and (f)), and occasionally identifies objects that were missed by OneFormer (image (e)).

\begin{figure*}[!htb]
   \centering
   \includegraphics[width=0.96\textwidth]{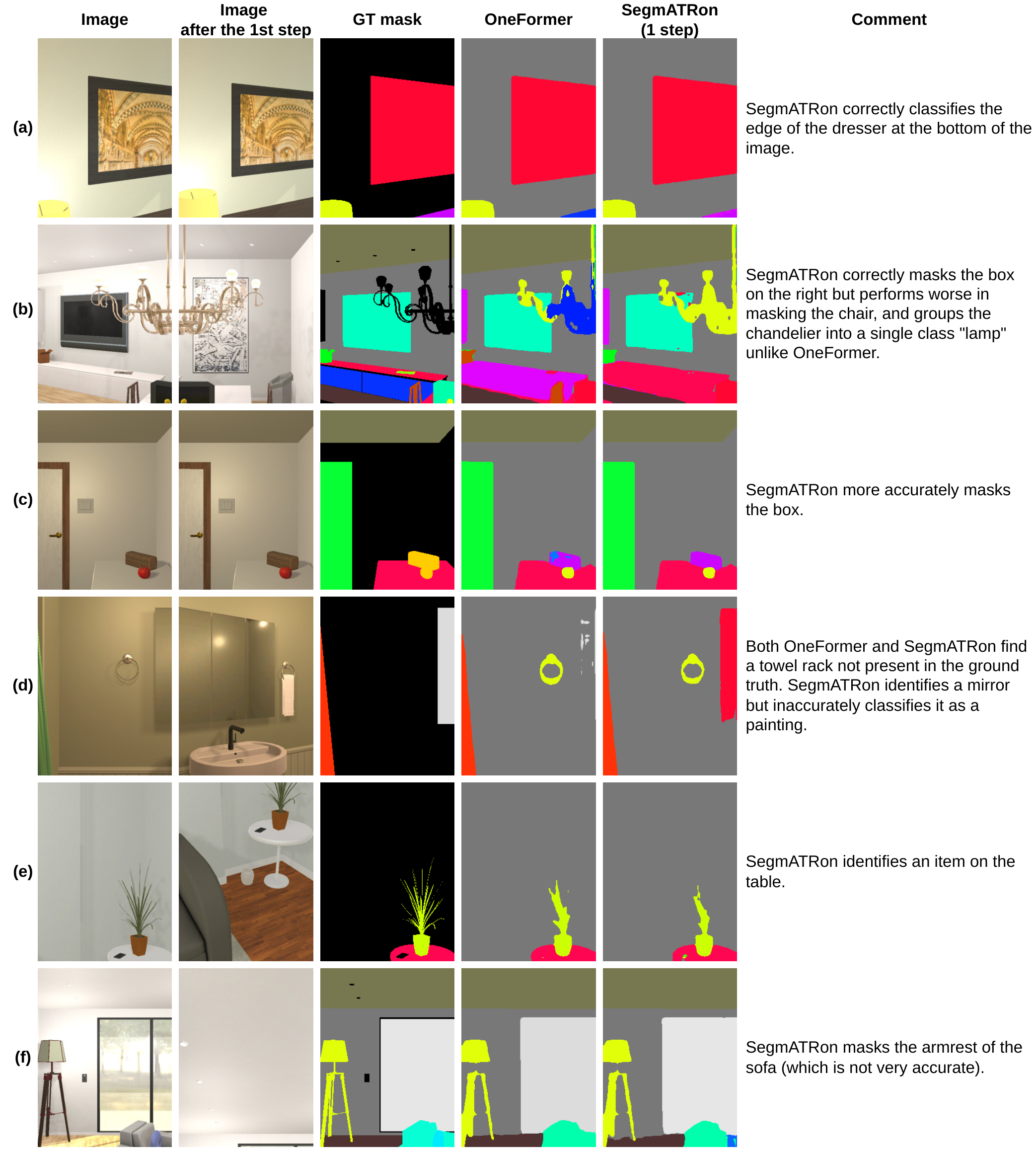}
   \caption{
  Visualized segmentation results on AI2-THOR validation set. The columns left-to-right refer to the input image, the image, received after one step, the ground truth, the outputs of the OneFormer model and our SegmATRon.
   }
   \label{fig:qualitative-results_ai2thor_supplementary}
\end{figure*}

Figure \ref{fig:qualitative-results_ai2thor_supplementary} shows more segmentation results of SegmATRon compared to Oneformer on the images rendered with AI2-THOR. Here, as observed in its results with Habitat, SegmATRon frequently exhibits more accurate classification of detected objects compared to OneFormer (image (a)). It also excels in discerning object masks (images (b)-(d)), albeit with occasional classification errors. Moreover, it identifies objects that were not detected by OneFormer (image (e)), although at times these may be extraneous objects (image (f)).

\end{document}